\documentclass[sigconf, table]{acmart}

\AtBeginDocument{%
  }

\copyrightyear{2025}
\acmYear{2025}
\setcopyright{acmlicensed}\acmConference[KDD '25]{Proceedings of the 31st ACM SIGKDD Conference on Knowledge Discovery and Data Mining V.2}{August 3--7, 2025}{Toronto, ON, Canada}
\acmBooktitle{Proceedings of the 31st ACM SIGKDD Conference on Knowledge Discovery and Data Mining V.2 (KDD '25), August 3--7, 2025, Toronto, ON, Canada}
\acmDOI{10.1145/3711896.3736991}
\acmISBN{979-8-4007-1454-2/2025/08}

\usepackage[utf8]{inputenc} 
\usepackage[T1]{fontenc}    
\usepackage{booktabs}       
\usepackage{amsfonts}       
\usepackage{nicefrac}       
\usepackage{microtype}      
\usepackage{xcolor}         
\usepackage{amsmath}
\usepackage{algorithm}
\usepackage{algorithmic}
\usepackage[most]{tcolorbox}
\usepackage{lipsum}
\usepackage{enumitem}
\usepackage{geometry}
\usepackage{multirow}
\usepackage{makecell}
\usepackage{tabularx}
\usepackage{caption}
\usepackage{wrapfig}
\usepackage{balance}
\usepackage{stfloats}
\usepackage{graphicx}
\usepackage{afterpage}
\usepackage{subcaption}
\usepackage{fancyhdr}

\newcommand{\modelname}{GraphLAMA }
\author{Junze Chen}
\affiliation{%
  \institution{Beijing University of Posts and Telecommunications}
  \city{Beijing}
  \country{China}
}
\email{junze@bupt.edu.cn}

\author{Cheng Yang}
\authornote{Corresponding author.}
\affiliation{%
  \institution{Beijing University of Posts and Telecommunications}
  \city{Beijing}
  \country{China}
}
\email{yangcheng@bupt.edu.cn}

\author{Shujie Li}
\affiliation{%
  \institution{Beijing University of Posts and Telecommunications}
  \city{Beijing}
  \country{China}
}
\email{shujieli@bupt.edu.cn}

\author{Zhiqiang Zhang}
\affiliation{%
  \institution{Ant Group}
  \city{Beijing}
  \country{China}
}
\email{lingyao.zzq@antfin.com}

\author{Yawen Li}
\affiliation{%
  \institution{Beijing University of Posts and Telecommunications}
  \city{Beijing}
  \country{China}
}
\email{warmly0716@126.com}

\author{Junping Du}
\affiliation{%
  \institution{Beijing University of Posts and Telecommunications}
  \city{Beijing}
  \country{China}
}
\email{junpingd@bupt.edu.cn}

\author{Chuan Shi}
\affiliation{%
  \institution{Beijing University of Posts and Telecommunications}
  \city{Beijing}
  \country{China}
}
\email{shichuan@bupt.edu.cn}
\renewcommand{\shortauthors}{Junze Chen et al.}
\begin{document}

\newpage
\setcounter{page}{1}
\title{GraphLAMA: Enabling Efficient Adaptation of Graph Language Models with Limited Annotations}

\renewcommand{\shortauthors}{Trovato et al.}

\begin{abstract}
  Large language models (LLMs) have demonstrated their strong capabilities in various domains, and have been recently integrated for graph analysis as graph language models (GLMs). With LLMs as the predictor, some GLMs can interpret unseen tasks described by natural language, and learn from a few examples in the prompts without parameter tuning, known as \textit{in-context learning (ICL)}. Another subset of GLMs utilizes abundant training labels to enhance model performance, known as \textit{instruction tuning}. 
  However, we argue that ICL on graphs has effectiveness issues due to fixed parameters and efficiency issues due to long context. Meanwhile, the large amount of labeled data required for instruction tuning can be difficult to obtain in real-world scenarios.
  To this end, we aim to introduce an extra parameter adaptation stage that can efficiently tailor GLMs to an unseen graph and task with only a few labeled examples, in exchange for better prediction accuracy and faster inference speed. For implementation, in this paper we propose \modelname method, with its model backbone and learning schemes specialized for efficient tuning and inference. Specifically, for the model backbone, we use a graph neural network (GNN) with several well-designed components (\textit{e.g.,} hop encodings, gating modules) to transform nodes into the representation space of LLM tokens. Task instructions can then be represented as a mixture of node and language tokens. In the pre-training stage, all model parameters except for the LLM will be trained with different tasks (\textit{i.e.,} node matching, node classification, and link prediction) to capture general knowledge. In the adaptation stage, only a few pre-trained parameters will be updated based on few-shot examples. Extensive experiments on few/zero-shot node classification and summary generation show that our proposed \modelname achieves state-of-the-art (SOTA) performance with 4.91\% absolute improvement in accuracy. Compared with ICL, our inference speed can be 10 times faster under 5-shot setting. Our code is available on GitHub at https://github.com/BUPT-GAMMA/GraphLAMA.
\end{abstract}



\maketitle

\section{Introduction}
Graph neural networks have become a powerful tool to understand and analyze graph-structured data \cite{dong2022edits, kipf2016semi}, facilitating progress in applications from various domains~\cite{lin2023comprehensive, lin2022effectively, wang2020traffic}. In real-world scenarios, graphs are usually accompanied by rich textual information as text-associated graphs (TAGs)~\cite{yang2021graphformers}, \textit{e.g.,} paper contents in citation networks~\cite{wen2023augmenting} and product descriptions in e-commerce networks~\cite{hu2020open}. 
Inspired by the recent advances in natural language processing,  efforts~\cite{zhang2024graphtranslator, tang2023graphgpt} have been made to integrate large language models into graph learning as graph language models~\cite{li2023survey} or graph foundation models (GFMs)~\cite{liu2023towards}. With LLMs as the predictor, some of these methods \cite{huang2024prodigy, liu2023one} can interpret unseen tasks described by natural language instructions, and learn from a few examples in the prompts without parameter tuning, known as \textit{in-context learning}~\cite{brown2020language}.  Meanwhile, some GLMs  \cite{tang2023graphgpt, tang2024higpt} use a large amount of labeled data for \textit{graph instruction tuning}~\cite{tang2023graphgpt} to adjust the model parameters.


Though ICL and instruction tuning have achieved significant results on graphs~\cite{tang2023graphgpt, huang2024prodigy, liu2023one}, these existing methods have the following limitations: 
(1) Bounded by fixed parameters, ICL-based methods have suboptimal performance~\cite{brown2020language, liu2022few} and limited improvement when we increase the number of examples in the prompt. Also, ICL has to include the information of all examples in the prompt, and thus suffer from extremely long context describing the topological structure and text content of nodes.
(2) While instruction tuning models can achieve excellent results, they require sufficient labeled data for training, which is difficult to obtain in real-world scenarios. The process of labeling data is both expensive and time-consuming, which highlights the importance of low-resource inference methods that require little or no labeled data~\cite{wen2023augmenting}.
\begin{table*}[ht]
\centering
\caption{Comparisons among different adaptation paradigms of GLMs. Existing methods primarily rely on in-context learning or instruction tuning. While using few-shot examples for efficient adaptation is common in GNNs, it has been largely overlooked in GLMs. Tuning with few-shot examples requires a separate model adjustment for each graph and task, necessitating a minimal number of tunable parameters, which presents significant challenges in backbone design. Also, how to effectively adapt GLMs with limited annotation remains an underdeveloped problem, making this issue far from trivial.}
\resizebox{\textwidth}{!}{
\footnotesize
\begin{tabular}{>{\centering\arraybackslash}m{3cm}|>{\centering\arraybackslash}m{3.2cm}|>{\centering\arraybackslash}m{3cm}|>{\centering\arraybackslash}m{3cm}}
\toprule
\textbf{} & \textbf{Graph In-context Learning} & \textbf{Graph Instruction Tuning} & \textbf{This Work} \\ 
\midrule
\textbf{Tuning data} & N/A & Large amount of data from various tasks & Few-shot examples for a specific task \\
\midrule
\textbf{Tuned parameters} & N/A & Relatively large & Relatively small \\
\midrule
\textbf{Tuning cost} & N/A & Relatively high & Relatively low \\
\midrule
\textbf{Graph and task-specific adaptation} & Yes & No & Yes \\
\midrule
\textbf{Inference cost} & Relatively high due to long context & Relatively low & Relatively low \\
\bottomrule
\end{tabular}
}
\label{tab:intro}
\end{table*}

To overcome the above limitations, we propose to introduce an extra parameter adaptation stage that efficiently tailors GLMs to each target graph and task with few-shot examples. Table~\ref{tab:intro} highlights the differences between this work and existing adaptation paradigms for GLMs, \textit{i.e.,} ICL and instruction tuning. For efficient adaptation, the number of tunable parameters needs to be minimized, which poses great challenges to the design of model backbone. Despite the recent progress of graph instruction tuning~\cite{tang2023graphgpt}, how to effectively adjust graph language models with only a few examples is still underdeveloped. 

In this paper, we propose \modelname with its model backbone and learning schemes specialized for efficient tuning and inference: (1) For backbone architecture, we first encode each target node and its neighborhood by a GNN, and augment the node embeddings with hop encoding. Then we transform node embeddings into the representation space of LLM tokens based on two gating modules and a projector. Respectively, the two gates extract task-invariant and task-related information from node embeddings. Finally, task instructions are represented as a mixture of node and language tokens. (2) In the pre-training stage, we train model parameters except the LLM with different tasks (\textit{i.e.,} node matching, node classification, and link prediction) to capture general knowledge. (3) In the adaptation stage, we use few-shot examples to update parameters in the task-related gate and hop encodings by a supervised loss. Here the number of tunable parameters is only $1/10^4$ of a 7B LLM, occupying merely 3MB storage space. (4) In the inference stage, we directly query the LLM without providing task examples in the prompt. Extensive experiments on few/zero-shot node classification and summary generation show that our \modelname achieves state-of-the-art (SOTA) performance with 4.91\% absolute improvement in accuracy. 

The contributions of this work are three-fold: (1)  We point out the limitations of current adaptation paradigms for GLMs, and propose to efficiently fine-tune GLMs for each target graph and task instead. (2) We propose \modelname with carefully-designed model backbone for implementation, enabling fast adaptation of GLM with only few-shot examples. (3) Extensive experimental results confirm the effectiveness and efficiency of GraphLAMA. The backbone components and pre-training tasks are also validated via ablation study.
\section{Related Works}
\label{related-works}
\textbf{Pre-training on Graphs.}
Self-supervised learning techniques, which design pretext tasks to utilize the inherent characteristics of graphs as extra guidance \cite{hu2020gpt,jing2021hdmi,xia2022simgrace}, have been widely explored for pre-training GNNs. These approaches can be further divided into contrastive and generative ones. Contrastive approaches emphasize learning representations by differentiating between positive and negative samples. Prominent methods in this category include GraphCL \cite{you2020graph} and GCA \cite{zhu2021graph}. Conversely, generative approaches concentrate on creating realistic samples that mimic the original graph data. Recent progress in this area includes methods like GraphMAE \cite{hou2022graphmae}, S2GAE \cite{tan2023s2gae} and AutoCF \cite{xia2023automated}, which reconstruct either masked features or edges as self-supervised learning tasks. In our method, we consider contrastive (\textit{i.e.,} node matching) and generative (\textit{i.e.,} link prediction) as well as supervised (\textit{i.e.,} classification) tasks during the pre-training stage, facilitating the learning of key information from text-attributed graphs.


\noindent\textbf{Prompt Tuning on Graphs.}
Inspired by advances in natural language processing, many researchers integrate prompt tuning techniques \cite{liu2023pre} into graph learning to enhance model transfer capabilities. 
Efforts such as GPPT \cite{sun2022gppt} and GraphPrompt \cite{liu2023graphprompt} focus primarily on closing the gap between pretext optimization and downstream predictions. These methods use prompts to enable knowledge transfer at various levels, including node, edge, and graph levels. GPF \cite{fang2024universal} and All-in-One \cite{sun2023all} incorporate diverse prompts into raw features or graph structures to accommodate a variety of tasks. Additionally, G2P2 \cite{wen2023augmenting} is specialized for classification tasks by employing graph-grounded prompting techniques to adapt the model to unseen datasets. 
However, these methods use GNNs as predictors, limiting them to predefined tasks and preventing them from following natural language instructions or offering explanations for their judgments.

\noindent\textbf{In-Context Learning on Graphs.}
Methods utilizing ICL for adaptation can be categorized into two types: those that use only LLM and those that combine GNN with LLM. The first type, such as NLGraph \cite{wang2024can} and LLMForGraph \cite{chen2024exploring}, directly employs LLMs to test on graph learning tasks. When utilizing ICL capabilities, they simply incorporate all available information of the examples into the input instructions. 
The second type combines the capabilities of GNNs and LLMs. OneForAll \cite{liu2023one} uses a GNN as the predictor and integrates language models to unify node features from different datasets. It remains restricted to predefined tasks and cannot provide explanations for its predictions.
An exception is Prodigy \cite{huang2024prodigy}, which designs an ICL graph to include task and data information, using a GNN to process it and make corresponding predictions. Like other GNN-based methods, it is confined to predefined tasks and lacks explainability.

\noindent\textbf{Graph Instruction Tuning.}
Some methods~\cite{tang2023graphgpt, zhang2024graphtranslator} use instruction tuning to enhance LLM performance on graph data. They employ GNNs to encode nodes and then align them with LLMs, followed by adapting LLMs to graph data through natural language graph task instructions. These methods perform well in experiments and retain the conversational abilities of large language models. However, these approaches have weak transfer capabilities and require a large amount of labeled data to construct sufficiently large graph task instructions for training, which is often impractical in real-world scenarios.
\section{Methodology}
\subsection{Problem Statement}
\noindent\textbf{Notations.} We focus on Text-Attributed Graphs (TAGs), where each node is associated with its own text content, and edges represent the relationships between nodes. We formally denote a TAG as $\mathcal{G}=\left(\mathcal{V}, \boldsymbol{A}, \boldsymbol{X}, \mathcal{S} \right)$, where $\mathcal{V}$ is the set of nodes, $\mathcal{S}$ is the set of node contents, $\boldsymbol{A} \in \{0, 1\}^{|\mathcal{V}| \times |\mathcal{V}|}$ is the adjacency matrix, and $\boldsymbol{X}$ denotes node features. We use subscripts to denote the information of specific nodes, \textit{e.g.,} the feature and content of node \(u\) are respectively represented as \( \boldsymbol{X}_u\) and \( \mathcal{S}_u\). Besides, we use superscripts on \( \boldsymbol{X} \) to indicate embeddings processed by a specific module. For example, \( \boldsymbol{X}_u^{\text{GNN}} \) denotes the embedding of node \( u \) after GNN processing. 


\noindent\textbf{Problem Definition.} For simplicity of notation, we formalize the problem with a single pre-training TAG $\mathcal{G}_{\text{pre}}$ and a target TAG $\mathcal{G}_{\text{target}}$ with different tasks. In our experiments, we have multiple datasets for pre-training. We denote the textual task descriptions for pre-training and adaptation as ${T}_{\text{task}}^{\text{pre}}$ and ${T}_{\text{task}}^{\text{target}}$, respectively. Pre-training tasks can be either supervised or self-supervised, while in the target task each node $v$ has a ground truth label $y_v$. Our goal is given task text ${T}_{\text{task}}^{\text{target}}$ and few-shot examples $\{(v_i, y_{v_i})\}_{i=1}^{K}$ on the target TAG, we aim to predict the labels of other nodes in $\mathcal{G}_{\text{target}}$.



\subsection{Framework Overview}
\label{sec:framework}

\begin{figure*}
    \centering
    \includegraphics[width=\linewidth]{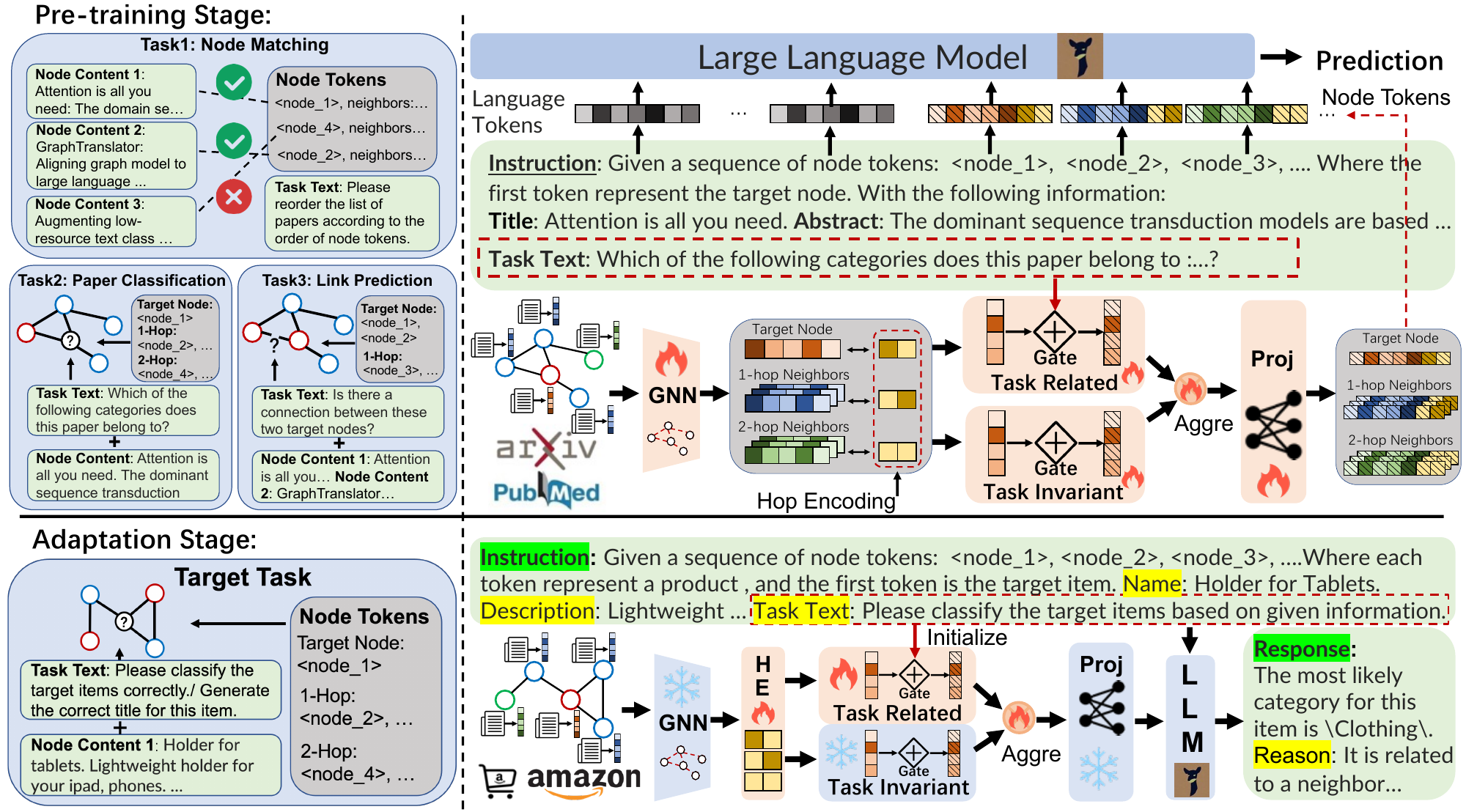}
    \caption{The overall framework of our proposed GraphLAMA, with explanations presented in Section~\ref{sec:framework}.}
    \label{fig:model}
\end{figure*}

The overall framework of our proposed \modelname{} is shown in Figure~\ref{fig:model}. Here each task query includes the tokens of target node as well as its neighborhood, the text content of target node, and the task text. The task query will be fed into an LLM for answering. Besides the LLM, our model backbone involves a GNN with hop encoding and gating modules to provide informative node token embeddings. In the pre-training stage, we train model parameters on $\mathcal{G}_{\text{pre}}$ with different tasks ${T}_{\text{task}}^{\text{pre}}$ (\textit{i.e.}, node matching, paper classification and link prediction) to capture general knowledge. In the adaptation stage, we use few-shot examples from the target TAG $\mathcal{G}_{\text{target}}$ to update a very small portion of pre-trained model parameters (\textit{i.e.}, the task-related gate and hop encodings) by a supervised loss. In the inference stage, we can directly query a target node without providing few-shot examples in the prompt.

\subsection{Backbone Architecture}
In our backbone, we first encode a target node and its neighborhood using a GNN. Then we augment the node embeddings with hop encoding, and transform them into the embedding space of language tokens based on two gating modules and a projector. Finally, the entire task query, including node tokens, target node content and task text, will be fed into an LLM.

\noindent\textbf{GNN.}
Given a TAG $\mathcal{G}=\left(\mathcal{V}, \boldsymbol{A}, \boldsymbol{X}, \mathcal{S}\right)$ with target node $v$, we  extract a neighborhood subgraph  $(\mathcal{N}(v), \boldsymbol{A}_{\mathcal{N}(v)}, \boldsymbol{X}_{\mathcal{N}(v)},  \mathcal{S}_{\mathcal{N}(v)})$, where $\mathcal{N}(v)$ represents the neighborhood nodes of 
$v$, and 
$\boldsymbol{A}_{\mathcal{N}(v)}$/ $\boldsymbol{X}_{\mathcal{N}(v)}$/ $\mathcal{S}_{\mathcal{N}(v)}$
represent the topology/features/contents of neighborhood subgraph. 
GNN performs neighbor aggregation for each node $\forall u \in\mathcal{N}(v)$, and can be represented as $g_{\theta}$, where $\theta$ denotes the set of trainable parameters. Formally, in the $l$-th layer of GNN, node's previous layer embedding $\boldsymbol{h}^{l-1}_u$ is combined with aggregated neighborhood vectors $\{\boldsymbol{h}^{l-1}_z, \forall z\in\mathcal{N}(u)\}$ using operation:
\begin{equation}
\begin{split}
    \label{eq1}
    \boldsymbol{h}_{u}^l = \sigma(\boldsymbol{W}_l \cdot {\rm CONCAT}(\boldsymbol{h}^{l-1}_u \cup {\rm AGGREGATE}_l \{\boldsymbol{h}^{l-1}_z, \forall z \in \mathcal{N}(u)\}).
\end{split}
\end{equation}
Finally, the GNN $g_{\boldsymbol{\theta}}(\cdot,\cdot)$ encodes the structure and feature information in the subgraph and yields node embeddings $\boldsymbol{X}_u^{\text{GNN}}=g_{\boldsymbol{\theta}}(\boldsymbol{A}_{\mathcal{N}(v)}, \boldsymbol{X}_{\mathcal{N}(v)})_u$, for all $u \in\mathcal{N}(v)$.

Due to the limited transfer capability inherent to the GNN itself, we train the GNN's encoding ability during the pre-training phase and then keep its parameters frozen during the adaptation stage. This configuration allows the GNN to focus solely on understanding the inherent features and the structure of the graph data.

\noindent\textbf{Hop Encoding.}
Next, hop encodings will be concatenated to node embeddings $\boldsymbol{X}_u^{\text{GNN}}$ for all $u \in\mathcal{N}(v)$. Specifically, hop encodings are fixed-length vectors, and set to be learnable parameters depending on the distance from the target node $v$. This effectively incorporates subgraph structure encoding into node embeddings, and helps highlight the differences between nodes at different distances. 

Formally, we denote hop encodings as $\{\boldsymbol{e}_i\}_{i=0}^{\lambda}$, where $\lambda$ is the maximum hop number in the extracted subgraph, and $\boldsymbol{e}_i$ is the encoding for the $i$-th hop neighbors. 
For a $i$-th hop neighbor node $u$, its embedding after hop encoding is given as:
\begin{equation}
    \label{eq2}
   \boldsymbol{X}_u^{\text{Hop}} = {\rm CONCAT}(\boldsymbol{X}_u^{\text{GNN}} \cup \boldsymbol{e}_i).
\end{equation}
Depending on the specific downstream task, the importance of neighbors from different hops may vary. Therefore, during the adaptation phase, we set hop encodings to be tunable. 

\noindent\textbf{Task-Invariant and Task-Related Gates.}
The Task-Invariant Gate (TIG) and Task-Related Gate (TRG) are both gating modules based on channel masking. TIG is to extract intrinsic characteristics independent of tasks from node embeddings, and will be frozen after the pre-training stage. In contrast, TRG aims to extract task-specific information from node embeddings based on task text. The learnable parameters in TRG are pre-trained and used for initialization in the adaptation stage. The embeddings after the two gates are then aggregated for further processing.



Formally, we denote the entire gating operation as $t_{\boldsymbol{\phi}}$ with $\boldsymbol{\phi}$ as the set of parameters. $\phi$ can be further divided into $\phi_{\text{inv}}$ in TIG, $\phi_{\text{rel}}$ in TRG, and $\phi_{\text{agg}}$ for aggregation. Here, $\phi_{\text{inv}}$ comprises a task-invariant weight matrix $\boldsymbol{W}_{\text{inv}}$ and a mask vector $\boldsymbol{m}_{\text{inv}}$. $\phi_{\text{rel}}$ includes a task-related weight matrix $\boldsymbol{W}_{\text{rel}}$. The task-related mask vector $\boldsymbol{m}_{\text{rel}}$ is parameterized by encoding task text $T_\text{task}$ with a small language model $f_{{\alpha}}$ such as Sentence-bert \cite{reimers2019sentence}. $\phi_{\text{agg}}$ contains an aggregation weight matrix $\boldsymbol{W}_{\text{agg}}$. 
The node embedding of node $u \in \mathcal{N}(v)$ after the gating modules can be defined as:
\begin{multline}
    \label{eq3}
    \boldsymbol{X}_u^{\text{Gate}} = \boldsymbol{W}_{\text{agg}} \cdot \rm CONCAT\{[(\boldsymbol{W}_{\text{rel}} \cdot \boldsymbol{X}_u^{\text{Hop}}) \odot \boldsymbol{m}_{\text{rel}}] \\ \cup [(\boldsymbol{W}_{\text{inv}} \cdot \boldsymbol{X}_u^{\text{Hop}}) \odot \boldsymbol{m}_{\text{inv}}]\}.
\end{multline}


This gating module is specialized for efficient adaptation, with fewer parameters than other encoding components, effectively reorganizing node features based on task text. During the adaptation phase, we treat  $\boldsymbol{m}_{\text{rel}}$ as tunable parameters initialized by $f_{\alpha}(T_\text{Task})$. Weight matrices $\boldsymbol{W}_{\text{rel}}$ and $\boldsymbol{W}_{\text{agg}}$ are also set as tunable, allowing the introduction of task-specific information to facilitate adaptation. 

\noindent\textbf{Projector.}
So far, there still exists a modality gap between the node embeddings and the tokens understandable by the LLM. Specifically, the dimensions and feature spaces of the node embeddings and the LLM tokens  differ from each other. To bridge this gap, we employ a projector $p_{\boldsymbol{\psi}}$ to map the embedding of node $u$ as 
\begin{equation}
    \label{eq5}
   \boldsymbol{X}^{\text{Proj}}_u = p_{\boldsymbol{\psi}}(\boldsymbol{X}^{\text{Gate}}_u).
\end{equation}
The projector has a relatively large number of parameters, so we freeze them after pre-training to perform translations.


\noindent\textbf{Large Language Model.} 
LLMs can function as either encoders or decoders. When used as an encoder, the LLM encodes node content into representations with rich semantic information. As a decoder, it can generate answers for a given task query by iteratively performing next token prediction.  

Formally, an LLM can be represented as $LLM_{\omega}$ where ${\omega}$ is the parameter pre-trained on broad language data. As an encoder, the LLM directly encodes the content $\mathcal{S}_u$ of node $u \in \mathcal{N}(v)$ as $\boldsymbol{C}_u = \text{LLM}_{{\omega}}(\mathcal{S}_u)$.
As a decoder, the LLM generates responses given node embeddings $\boldsymbol{X}_{\mathcal{N}(v)}^{\text{Proj}}$, target node content $\mathcal{S}_v$ and task text ${T}_{\text{task}}$:
\begin{equation}
    \label{eq7}
    \hat{y}_{v} =\text{LLM}_{{\omega}}[\boldsymbol{X}_{\mathcal{N}(v)}^{\text{Proj}}, \mathcal{S}_v, {T}_{\text{task}}].
\end{equation}
    We utilize the Vicuna model~\cite{chiang2023vicuna}, an open-source chatbot developed by fine-tuning Llama~\cite{touvron2023llama} with conversations shared by users. To avoid the problems of catastrophic forgetting and the prohibitive training expenses due to managing a massive parameter set, we maintain the LLM parameters $\omega$ as fixed during all stages. 

\subsection{Pre-training Stage}
In the pre-training stage, we consider three tasks to ensure that the main components are adequately trained and can learn general information from the TAG data. The tasks include node matching (Task 1), paper classification (Task 2), and link prediction (Task 3). During this stage, all model parameters in the framework, except for the large language model, will be updated by gradient descent.

\noindent\textbf{Node Matching.}
Following previous researches  \cite{tang2023graphgpt,zhang2024graphtranslator}, we start pre-training by an unsupervised contrastive learning task to train the model's encoding and alignment abilities. In implementation, we align node embeddings $\boldsymbol{X}_{\mathcal{N}(v)}^{\text{Proj}}$ with their corresponding text encodings $\boldsymbol{C}_{\mathcal{N}(v)}$. Let $ \hat{\boldsymbol{X}}_{\mathcal{N}(v)} = \text{norm}(\boldsymbol{X}_{\mathcal{N}(v)}^{\text{Proj}}) $ and $ \hat{\boldsymbol{C}}_{\mathcal{N}(v)} = \text{norm}(\boldsymbol{C}_{\mathcal{N}(v)}) $ represent the normalized node and text embeddings respectively. The loss function for alignment computed as:
\begin{equation}
    \label{eq8}
    \mathcal{L} = \frac{1}{\left | \mathcal{N}(v) \right |} \sum_{u \in \mathcal{N}(v)} \exp\left(-\gamma \cdot \cos(\hat{\boldsymbol{X}}_u, \hat{\boldsymbol{C}}_u)\right) +  \|\hat{\boldsymbol{X}}_u - \hat{\boldsymbol{C}}_u\|^2,
\end{equation}
where $\gamma$ is fixed as $0.01$, $\cos(\cdot, \cdot)$ is the cosine similarity, and the last term is the square of L2-norm.

\noindent\textbf{Paper classification \& Link Prediction.}
To further equip the model with the capability to address practical tasks, we also employ two supervised tasks, paper classification and link prediction, during the pre-training phase. We combine node tokens with language tokens as instructions for the LLM. The instruction examples for the two tasks are shown below:
\begin{tcolorbox}[colback=gray!10, colframe=black, rounded corners, boxrule=1.5pt, fontupper=\normalsize, left=2mm, right=2mm, top=1mm, bottom=1mm]
    \textbf{Task2: Paper Classification} \newline
    \textbf{Instruction:} Given a sequence of node tokens: <node\_1>, <node\_2>, <node\_3>... 
    Where the first token represents the target node. With the following information:\newline
    \textit{Title:} Attention is all you need. 
    \textit{Abstract:} The dominant sequence transduction models are ... \newline
    \textit{Task Text:} Which of the following categories does this paper belong to? 
\end{tcolorbox}
\begin{tcolorbox}[colback=gray!10, colframe=black, rounded corners, boxrule=1.5pt, fontupper=\normalsize, left=2mm, right=2mm, top=1mm, bottom=1mm]
    \textbf{Task3: Link Prediction} \newline
    \textbf{Instruction:} Given two sequences of node tokens: <node\_1>, <node\_2>, <node\_3>... and <node\_i>, <node\_(i+1)>, <node\_(i+2)>...
    Where the first token and the i-th token represent the target nodes. With the following information:\newline
    \textit{Title1:} Attention is all you need.
    \textit{Abstract1:} The dominant sequence transduction ... \newline
    \textit{Title2:} GraphTranslator: Aligning Graph Model ...
    \textit{Abstract2:} Large language models  ... \newline
    \textit{Task Text:} Is there a connection between these two target nodes?
\end{tcolorbox}
Specifically, we obtain LLM predictions $\hat{y}$ by Eq.~(\ref{eq7}), and compute the cross-entropy loss with ground truth label $y$ for each task.

\subsection{Adaptation Stage}
In this stage, we conduct fine-tuning with few-shot examples of target task and TAG. Here most model parameters are frozen, and the tunable parameters only involve $\boldsymbol{m}_{\text{rel}}$, $\boldsymbol{W}_{\text{rel}}$ and $\boldsymbol{W}_{\text{agg}}$, thus enabling a fast adaptation to the downstream task. The loss function for parameter fine-tuning is also cross entropy, the same as the supervised pre-training tasks. For each target task, we need to store a specific set of $\boldsymbol{m}_{\text{rel}}$, $\boldsymbol{W}_{\text{rel}}$ and $\boldsymbol{W}_{\text{agg}}$, occupying less than $3$MB space in practice. Note that the target data domain and task may significantly vary from the pre-training ones. For example, an instruction example of target task in our experiments is as follow:
\begin{tcolorbox}[colback=gray!10, colframe=black, rounded corners, boxrule=1.5pt, fontupper=\normalsize, left=2mm, right=2mm, top=1mm, bottom=1mm]
    \textbf{Target Task Example} \newline
    \textbf{Instruction:} Given a sequence of node tokens: <node\_1>, <node\_2>, <node\_3>... 
    Where the first token represents the target node. With the following information:\newline
    \textit{Description:} Lightweight holder for your ipad, phones. Product Dimensions: 3.3 x 8.6 ... \newline
    \textit{Task Text:} Generate a suitable title for this product? 
\end{tcolorbox}

\noindent\textbf{Discussion.} In the inference stage, the task query has the same format with that in the adaptation stage. Compared with previous methods, our advantages are two-fold: 
(1) Compared to instruction tuning, our method only requires a few-shot number of data to achieve highly effective transfer. The well-designed backbone  
maximizes \modelname's inference capability on unseen datasets and unseen tasks. In contrast, the instruction tuning method requires abundant training labels.
(2) Compared to ICL, the model parameters in our methods are adjusted for the target task and TAG, and thus can better adapt to downstream scenarios. In contrast, increasing the number of examples for ICL only brings marginal performance improvements while significantly reducing the model's inference efficiency. We name our method as \modelname (\textbf{Graph} \textbf{LA}nguage \textbf{M}odel \textbf{A}dapter), and present the pseudo code in Appendix~\ref{pseudo-code}.

\section{Experiments}

In this section, we conduct extensive experiments to validate the effectiveness of our proposed method from multiple perspectives. We aim to answer the following research questions (\textbf{RQ}s): \textbf{RQ1:} How effective is our proposed \modelname in learning scenarios with limited annotations? \textbf{RQ2:} 
What is the rationale behind our backbone design and choice of pre-training tasks?
\textbf{RQ3:} What are the details of \modelname's efficiency at each stage? \textbf{RQ4:} Do the LLM-generated predictions in \modelname offer some interpretability?
{
\renewcommand{\arraystretch}{1.5}
\begin{table*}[ht]
\centering
\caption{Results on few-shot classification, with the highest results displayed in bold and the second highest underlined. The dash (-) indicates the text input exceeded the length limits. The tag (Tuned) means that we have fine-tuned the model's parameters using the corresponding few-shot data, while the tag (ICL) indicates that we included the corresponding few-shot data in the input for in-context learning.}
\normalsize
\setlength{\tabcolsep}{1pt}
\resizebox{\textwidth}{!}{
\begin{tabular}{@{} c|c|c>{\columncolor{gray!30}}cc>{\columncolor{gray!30}}cc>{\columncolor{gray!30}}c|c>{\columncolor{gray!30}}cc>{\columncolor{gray!30}}cc>{\columncolor{gray!30}}c|c @{}}
\Xhline{1.5pt}
\multirow{2}{*}{Dataset}     & \multirow{2}{*}{Shots}    & \multirow{2}{*}{\makecell{GraphMAE \\ (Tuned)}} & \multirow{2}{*}{\makecell{All-In-One \\ (Tuned)}} & \multirow{2}{*}{\makecell{GraphPrompt \\ (Tuned)}} & \multirow{2}{*}{\makecell{Prodigy \\ (ICL)}} & \multirow{2}{*}{\makecell{OneForAll \\ (ICL)}}  & \multirow{2}{*}{\makecell{G2P2 \\ (Tuned)}} & \multirow{2}{*}{\makecell{Vicuna \\ (ICL)}} & \multirow{2}{*}{\makecell{NLGraph \\ (ICL)}} & \multirow{2}{*}{\makecell{LLaGA \\ (Tuned)}} & \multirow{2}{*}{\makecell{G-Retriever \\ (Tuned)}} & \multirow{2}{*}{\makecell{GraphGPT \\ (ICL)}} & \multirow{2}{*}{\makecell{GraphGPT \\ (Tuned)}} & \multirow{2}{*}{\makecell{\modelname \\ (Tuned)}} \\ 
& & & \raisebox{1ex}{(Tuned)} & &\raisebox{1ex}{(ICL)} & & \raisebox{1ex}{(Tuned)}& &\raisebox{1ex}{(ICL)} & &\raisebox{1ex}{(Tuned)} & &\raisebox{1ex}{(Tuned)} & \\
\hline 
\multirow{3}{*}{\makecell{Cora \\ 5ways}}  & 5  & 27.80±1.02 & 40.56±1.32 & 36.12±2.01 & 52.94±0.72 & 50.54±1.36 & \underline{74.52±1.77} & 48.50±1.23 & 47.71±1.5 & 68.68±2.56 & 50.47±1.17 & 51.63±2.44 & 49.96±1.64 & \textbf{74.54±1.61} \\
            & 20 & 28.08±1.33 & 41.57±0.70 & 36.95±1.77 & 53.51±1.21 & 57.32±2.78  & \underline{78.69±1.48} & -           & -       & 66.14±1.74   & 50.86±1.10 &51.09±3.81 & 48.20±1.94 & \textbf{80.56±2.11} \\
            & 50 & 28.88±1.02 & 42.24±0.83 & 37.65±1.74 & 54.51±1.49 & 56.36±2.21  & \underline{81.76±1.56} & -           & -       & 67.09±2.55   & 50.36±2.09       &52.40±3.02 & 48.40±2.75 & \textbf{88.96±1.93} \\
\hline
\multirow{3}{*}{\makecell{Cora \\ 10ways}} & 5  & 17.99±0.64 & 30.07±1.30 & 26.22±1.69 & 44.09±0.88 & 44.50±1.34  & \underline{55.61±2.04} & 35.16±2.03 & 36.83±2.67 & 47.51±1.53 & 39.15±2.20 & 39.22±2.64 & 38.97±1.58 & \textbf{58.61±1.92} \\
            & 20 & 16.04±0.83 & 31.52±1.01 & 27.09±2.08 & 46.74±1.24 & 47.08±1.18  & \underline{58.29±2.37} & -           & -       & 48.95±3.08    & 39.28±2.11      &39.32±4.08 & 39.04±1.25 & \textbf{69.04±1.69} \\
            & 50 & 18.54±0.64 & 28.21±5.07 & 27.64±2.31 & 48.28±0.90 & 48.62±1.41  & \underline{60.50±2.74} & -           & -       & 47.91±3.25    & 39.40±2.17      &40.36±2.30 & 38.81±2.74 & \textbf{79.71±1.84} \\
\hline
\multirow{3}{*}{\makecell{Wiki-CS \\ 5ways}} &5 & 27.74±0.61 & 45.65±1.83 & 34.25±1.86 & 50.36±0.72 & 50.76±1.33  & \textbf{71.37±2.18} & 50.19±1.57 & 45.74±2.76 & 62.71±2.85 & 54.04±2.91 &56.80±1.82 & 54.26±2.45 & \underline{70.18±2.07} \\
            & 20 & 29.42±0.57 & 40.44±1.59 & 34.37±1.71 & 51.60±0.61 & 55.27±1.06  & \underline{74.41±1.58} & -           & -      & 61.45±2.66  & 55.39±2.41 & 56.01±1.55 & 54.78±2.43 & \textbf{76.88±2.30} \\
            & 50 & 29.79±0.94 & 46.60±1.20 & 35.16±1.79 & 53.61±0.72 & 56.22±0.71  & \underline{76.21±2.50} & -           & -      & 62.13±2.74  & 55.13±1.20   & 57.22±1.79 & 55.72±2.11 & \textbf{81.99±1.64} \\
\hline
\multirow{3}{*}{\makecell{Wiki-CS \\ 10ways}} & 5  & 19.29±1.02 & 32.74±2.89 & 29.13±2.47 & 50.37±1.15 & 48.65±0.55 & \textbf{63.38±2.36} & 45.07±2.56 & 44.73±1.58 & 56.31±1.80 & 50.78±2.09 & 49.73±1.66 & 49.36±1.02 & \underline{62.24±2.18} \\
            & 20 & 19.25±1.29 & 36.83±1.85 & 29.59±2.20 & 51.36±0.91 & 50.43±0.82  & \underline{65.07±2.48} & -           & -     & 58.38±2.95   & 50.25±2.06   & 49.35±2.35 & 48.97±2.92 & \textbf{68.20±2.40} \\
            & 50 & 19.28±0.87 & 37.40±1.85 & 29.76±2.16 & 53.37±1.01 & 52.66±1.32  & \underline{68.36±1.53} & -           & -     & 57.53±2.71   & 50.69±2.84   & 49.20±1.54 & 48.89±2.77 & \textbf{75.81±2.34} \\
\hline
\multirow{3}{*}{\makecell{Products \\ 5ways}} & 5  & 19.62±1.14 & 31.46±2.84 & 32.08±1.66 & 47.58±0.59 & 46.72±0.77 & \underline{68.22±1.25} & 44.44±1.97 & 45.80±2.12 & 55.43±1.84 & 49.13±1.32 & 50.72±2.29 & 48.50±2.33 & \textbf{69.67±1.64} \\
               & 20 & 23.33±0.58 & 31.93±3.14 & 32.81±2.21 & 50.67±1.07 & 48.59±0.77 & \underline{70.87±1.11} & -          &  -         & 55.29±2.99  & 49.80±1.26 & 50.78±1.94 & 48.64±2.29 & \textbf{72.61±2.19} \\
               & 50 & 18.81±0.82 & 32.11±2.16 & 32.86±2.08 & 51.94±0.90 & 49.16±1.04 & \underline{82.62±0.63} & -          & -          & 55.12±3.08  & 49.05±1.77 & 50.30±1.81 & 48.58±1.16 & \textbf{84.82±1.58} \\
\hline
\multirow{3}{*}{\makecell{Products \\ 10ways}} & 5  & 12.88±1.14 & 20.30±8.65 & 26.30±2.42 & 40.48±0.53 & 39.03±0.65 & \underline{53.83±1.67} & 39.44±1.87 & 36.45±1.66 & 48.33±2.37 & 41.33±1.69 & 36.94±2.01 & 35.06±2.86  & \textbf{58.30±2.17} \\
                & 20 & 11.47±1.47 & 19.07±6.11 & 26.77±2.29 & 41.77±0.99 & 40.51±0.69 & \underline{55.79±0.52} & -          & -          & 49.22±1.14  & 41.79±2.80       &36.53±1.87 & 35.35±2.56 & \textbf{64.85±1.86} \\
                & 50 & 13.82±1.36 & 22.49±3.79 & 26.96±2.02 & 42.30±0.74 & 42.46±0.78 & \underline{64.92±0.67} & -         & -          & 49.24±2.93   & 41.66±2.33      & 36.92±2.42 & 35.21±2.20 & \textbf{75.81±1.52} \\     
\hline
\end{tabular}
}
\label{fewshots}
\end{table*}
}

\subsection{Experimental Setup}

\noindent\textbf{Datasets.} 
Following previous works \cite{wen2023augmenting,tang2023graphgpt}, we pre-train on  ArXiv and PubMed datasets, and test on Cora, Wiki-CS, and ogbn-products datasets. More dataset details can be found in Appendix~\ref{datasets-details}.

\noindent\textbf{Evaluation.}
We consider two different tasks to demonstrate our model's capabilities, encompassing classification and summary. To simulate real-world scenarios with limited labels, we adopt the classic setting of few-shot learning. For the classification task, we randomly select \(K\) samples per class as the training data during the adaptation stage. We construct six different classification tasks using three test datasets, namely Cora-5ways, Cora-10ways, Wiki-CS-5ways, Wiki-CS-10ways, Products-5ways, and Products-10ways. The first part of each task name indicates the dataset used, while the second part specifies the number of classes. We use accuracy as the evaluation metric for both few-shot and zero-shot classification tasks, and calculate the standard deviation based on different random seeds. For each classification task, we follow the existing settings and use 100 target nodes per way as testing data. For summary generation task, we utilize the ``title'' field from node's text content as ground truth label, as the title can be considered as a summary of the entry's information. We also collect $K$ samples per class as tuning examples and conduct tests on all three test datasets. The summary generation task uses BLEU-1 score as the evaluation metric.
We select a number of SOTA methods as baselines for comparison, baseline information can be found in~\ref{baseline-details}. The implementation details of our model can be found in Appendix~\ref{implementation-details}.

\subsection{Few-shot Experiments (RQ1)}
\begin{figure*}[!th]
    \centering
    \includegraphics[width=\linewidth]{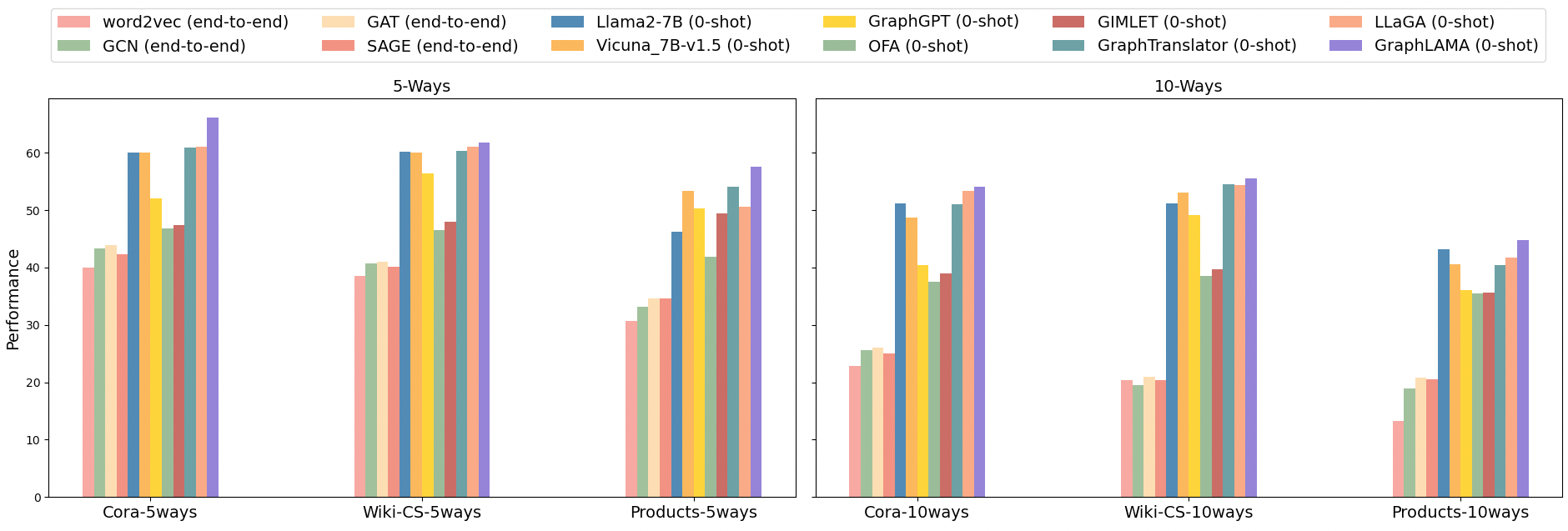}
    \caption{Results on zero-shot classification. Our proposed \modelname always has the highest accuracy.}
    \label{fig:0-shot}
\end{figure*}

\begin{figure*}[!htb]
\centering
\begin{minipage}{0.8\linewidth}
    \centering
    \includegraphics[width=\linewidth]{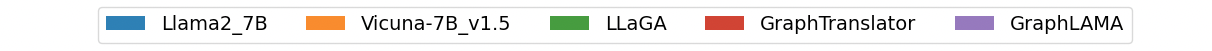} 
\end{minipage}

\makebox[\textwidth]{%
    \begin{minipage}{0.34\linewidth}
        \centering
        \includegraphics[width=\linewidth]{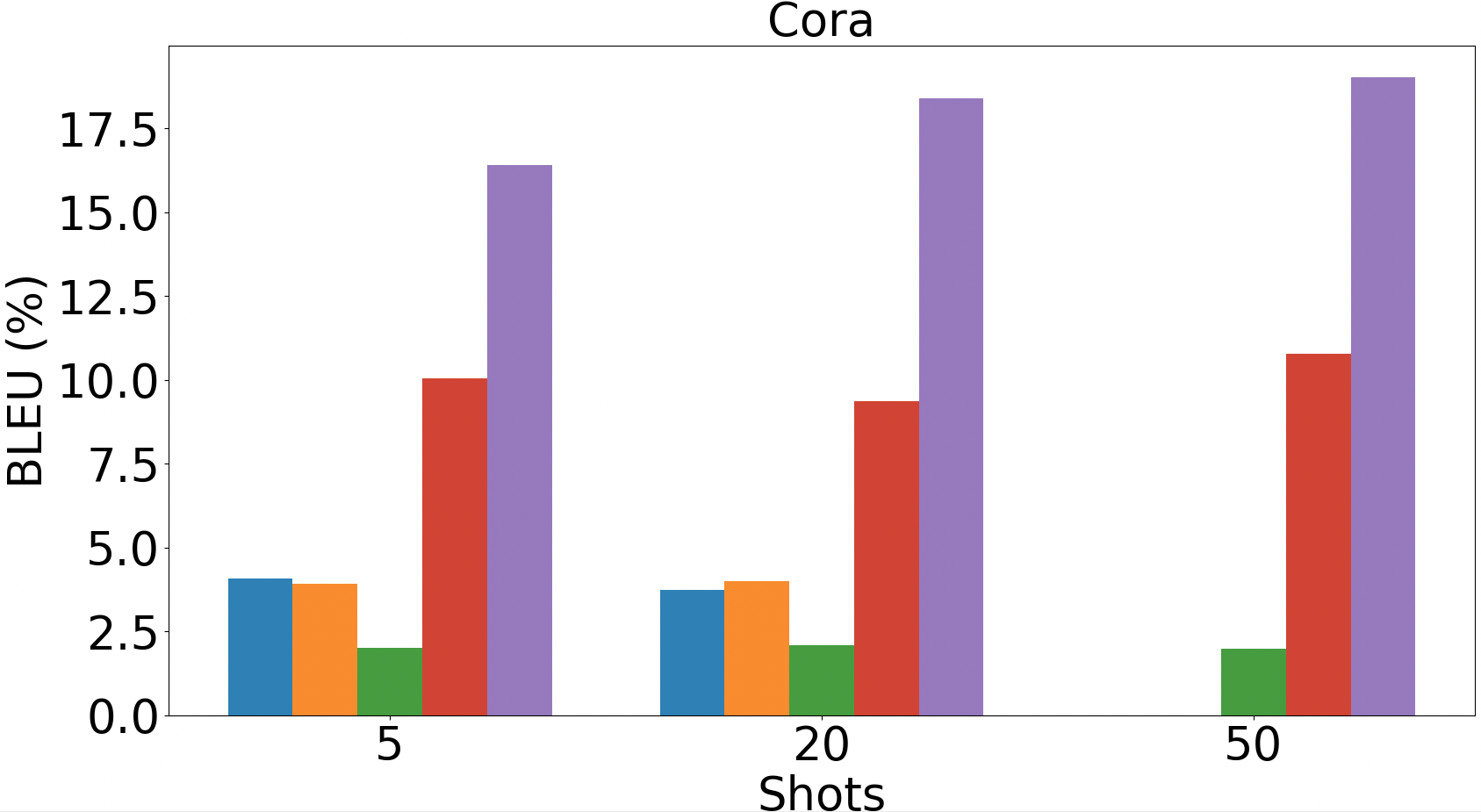}
    \end{minipage}
    \hspace{-0.1cm}
    \begin{minipage}{0.34\linewidth}
        \centering
        \includegraphics[width=\linewidth]{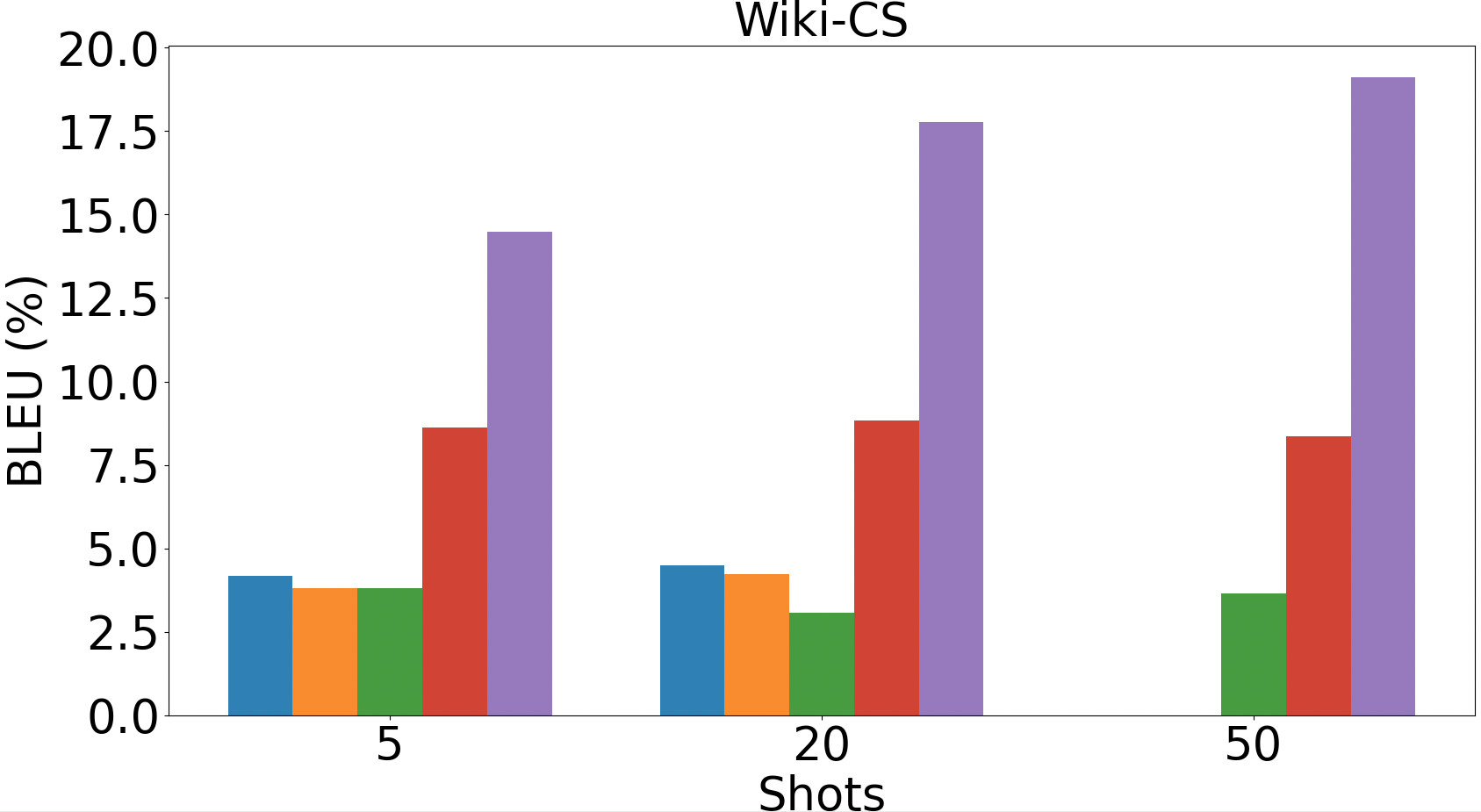} 
    \end{minipage}
    \hspace{-0.1cm}
    \begin{minipage}{0.33\linewidth}
        \centering
        \includegraphics[width=\linewidth]{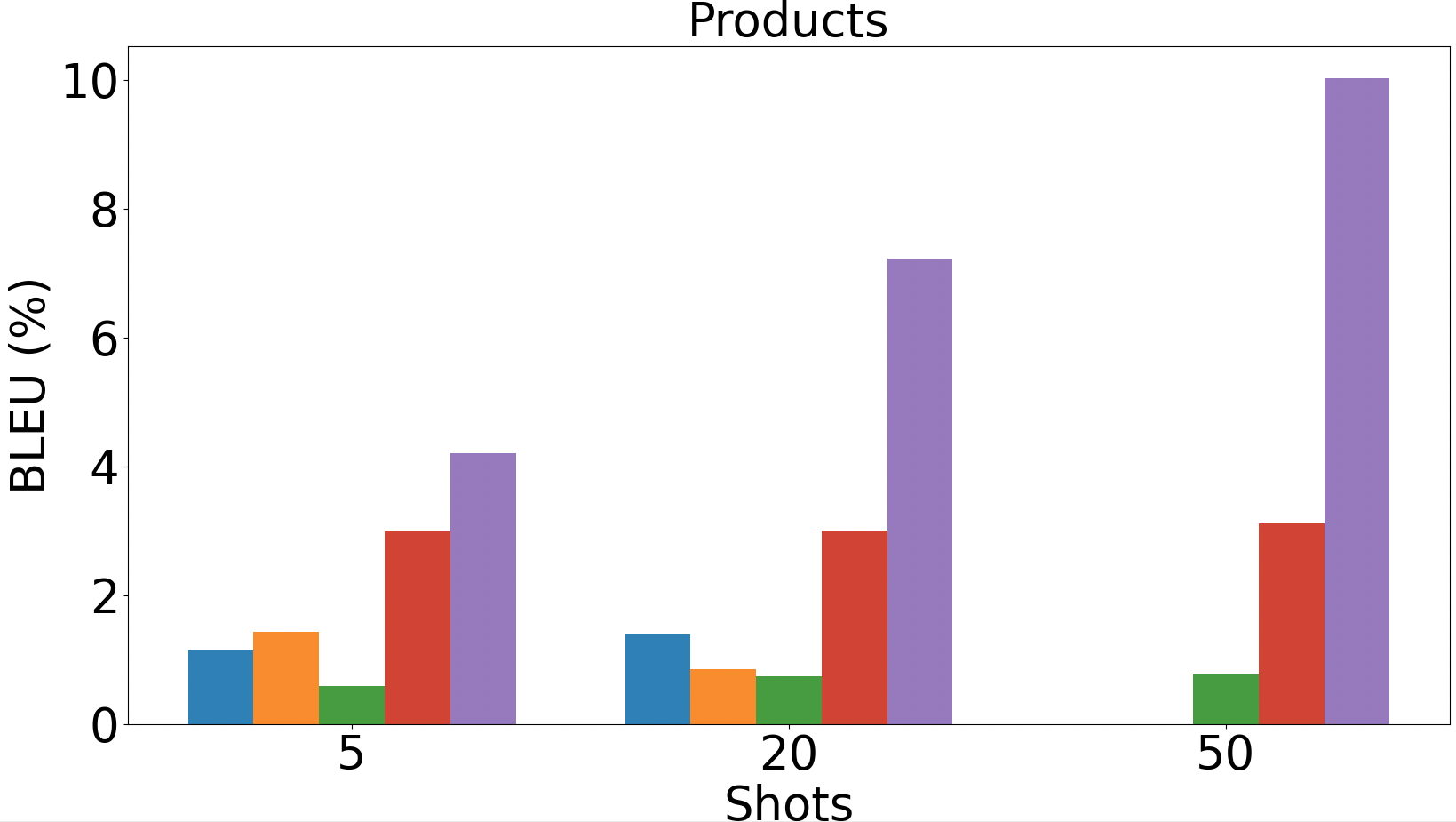} 
    \end{minipage}
}
\caption{The results of summary generation task in three test datasets. The missing result in 50-shot setting is because the text input exceeds the length limit.}
\label{fig:summary}
\end{figure*}

To answer \textbf{RQ1}, we conduct few-shot classification experiments under six settings, and the results are shown in Table~\ref{fewshots}. Additionally, we conduct zero-shot classification experiments under the same settings, with results shown in Figure~\ref{fig:0-shot}. Furthermore, we conduct few-shot summary generation task as shown in Figure~\ref{fig:summary}.
The results demonstrate that: (1) Our \modelname method consistently outperforms SOTA methods in few-shot classification, always delivering the highest or second-highest accuracy in its predictions. On average, \modelname has 4.91\% absolute improvement in classification accuracy against the best baseline. This indicates that our adaptation strategy provides the model with a strong transfer capability. 
(2) Our approach demonstrates high accuracy even with just 5-shot examples and shows considerable improvement as the number of shots increases. In contrast, the performance of baseline methods shows only marginal improvement as the number of shots increases. 
(3) \modelname also exhibits high performance on the products dataset, whereas other baselines are significantly affected. The textual content of the products dataset differs greatly from that in the pre-training academic citation network, making it a challenging task in our experiments. However, our model is only minimally impacted, demonstrating our method's capability to handle cross-domain data.
(4) Notably, GraphMAE, All-In-One, and GraphPrompt do not support textual input, resulting in lower performance due to the lack of one input modality. Conversely, methods using GNN as a predictor excel in classification tasks but cannot perform the summary generation task, aligning with their typical characteristics. Besides, experiments show that the performance of the GraphGPT (ICL) version does not significantly improve with an increased number of shots. On the other hand, GraphGPT (Tuned) version is also limited in performance due to the large number of 
tunable parameters in the backbone and the lack of design for few-shot transfer.

From the zero-shot experimental results, we can see that our proposed \modelname demonstrates the best performance in all zero-shot tasks with 3.10\% absolute improvement in accuracy, indicating that our backbone architecture successfully integrates text and graph information. Notably, some baseline models were not specifically designed for 0-shot scenarios, and the lack of appropriate transfer training settings during the experiments introduced some noise into the information, which explains why the original Llama model performed relatively better.

From the summary generation task, it is evident that by using LLMs as predictors, our model can handle a wider range of tasks through generative capabilities. It is noteworthy that G2P2 model is specifically tailored for classification tasks, where it achieves the second-best performance, but it cannot handle other tasks like summary generation. 
We further equip GraphGPT with Vicuna-13B-v1.5 and LLaMA-8B-v3 to compare our method with a direct scaling of GLMs in terms of model size and training data. The corresponding results are provided in Appendix~\ref{scaling}. The results indicate that \modelname has a powerful transfer capability.


\subsection{Ablation Study (RQ2)}
To address RQ2, we conduct a series of ablation studies to examine the effectiveness and necessity of the backbone design and the pre-training tasks in \modelname.

\noindent\textbf{Ablation Study for Adaptation Stage.}
Firstly, we compare our proposed adaptation stage with existing lightweight tuning method, with results shown in Table~\ref{ablation_adaptation}. Specifically, GraphLAMA (LoRA) tunes LLM via LoRA, and \modelname (Original) represents our proposed adaptation approach.  As shown, our method consistently outperforms across all settings with a 26.7\% absolute improvement on averge. A possible reason for the underperformance of GraphLAMA (LoRA) is that the number of trainable parameters in LoRA is still ten times larger than that in our method. Under such limited annotations, LoRA fails to achieve effective adaptation.

\begin{center}
{
\renewcommand{\arraystretch}{1.1}
\captionof{table}{Ablation for adaptation stage.}
\resizebox{0.45\textwidth}{!}{
\begin{tabular}{@{}c|c|>{\columncolor{gray!30}}c|c@{}}
\Xhline{1.5pt}
Dataset & Shots & \makecell{\hspace{2pt}\modelname (LoRA)\hspace{2pt}} & \makecell{\hspace{2pt}\modelname (Original)\hspace{5pt}} \\
\hline
\multirow{3}{*}{\makecell{Cora-5ways}}
            & 5  & 48.72±1.87 & \textbf{74.54±1.61} \\
            & 20 & 48.55±2.14 & \textbf{80.56±2.11} \\
            & 50 & 49.13±1.36 & \textbf{88.96±1.93} \\
\hline
\multirow{3}{*}{\makecell{Cora-10ways}}
            & 5  & 39.19±2.32 & \textbf{58.61±1.92} \\
            & 20 & 39.31±1.54 & \textbf{69.04±1.69} \\
            & 50 & 39.27±2.01 & \textbf{79.71±1.84} \\
\hline
\multirow{3}{*}{\makecell{Wiki-CS-5ways}}
            & 5  & 55.38±1.77 & \textbf{70.18±2.07} \\
            & 20 & 55.39±2.24 & \textbf{76.88±2.30} \\
            & 50 & 55.44±1.68 & \textbf{81.99±1.64} \\
\hline
\multirow{3}{*}{\makecell{Wiki-CS-10ways}}
            & 5  & 49.63±1.95 & \textbf{62.24±2.18} \\
            & 20 & 49.14±2.11 & \textbf{68.20±2.40} \\
            & 50 & 49.21±1.42 & \textbf{75.81±2.34} \\
\hline
\multirow{3}{*}{\makecell{Products-5ways}}
            & 5  & 48.78±1.61 & \textbf{69.67±1.64} \\
            & 20 & 48.82±2.09 & \textbf{72.61±2.19} \\
            & 50 & 49.10±1.83 & \textbf{84.82±1.58} \\
\hline
\multirow{3}{*}{\makecell{Products-10ways}}
                & 5  & 35.48±1.56 & \textbf{58.30±2.17} \\
                & 20 & 35.90±2.47 & \textbf{64.85±1.86} \\
                & 50 & 35.73±1.29 & \textbf{75.81±1.52} \\     
\hline
\end{tabular}
}
\label{ablation_adaptation}
}
\end{center}
{
\renewcommand{\arraystretch}{1.05}
\small
\setlength{\tabcolsep}{1pt}
\captionof{table}{Ablation for backbone component.}
\resizebox{0.48\textwidth}{!}{
\begin{tabular}{@{}c|c|>{\columncolor{gray!30}}c|c|>{\columncolor{gray!30}}c|c@{}}
\Xhline{1.5pt}
Dataset     & Shots    &   \makecell{w/o \\ Gates Module} & \makecell{w/o \\ Hop Encoding} & \makecell{w/o \\ Task-Related Mask} & \modelname \\ 
\hline
\multirow{4}{*}{\makecell{Cora \\ 5ways}}
            & 0   & 49.34±2.49 & 60.00±1.24 & 55.13±2.01 & \textbf{66.11±1.32} \\
            & 5  & 51.11±2.39 & 68.46±0.86 & 62.10±0.95 & \textbf{74.54±1.61} \\
            & 20 & 48.04±1.89 & 74.44±1.40 & 74.17±0.51 & \textbf{80.56±2.11} \\
            & 50 & 50.10±1.80 & 75.52±0.82 & 86.80±0.77 & \textbf{88.96±1.93} \\
\hline
\multirow{4}{*}{\makecell{Cora \\ 10ways}} 
            & 0   & 30.80±1.55 & 45.88±1.44 & 21.79±1.51 & \textbf{54.01±1.09} \\
            & 5  & 29.24±1.90 & 52.55±0.82 & 34.36±0.71 & \textbf{58.61±1.92} \\
            & 20 & 30.60±2.15 & 53.67±0.69 & 42.58±0.37 & \textbf{69.04±1.69} \\
            & 50 & 29.39±1.62 & 54.25±1.41 & 56.00±1.24 & \textbf{79.71±1.84} \\
\hline
\multirow{4}{*}{\makecell{Wiki-CS \\ 5ways}} 
            & 0   & 57.37±2.40 & 55.21±1.28 & 59.03±1.84 & \textbf{61.74±1.27} \\
            & 5  & 56.20±2.10 & 56.55±1.48 & 63.60±1.27 & \textbf{70.18±2.07} \\
            & 20 & 57.69±1.65 & 70.10±1.18 & 70.00±0.68 & \textbf{76.88±2.30} \\
            & 50 & 57.65±2.39 & 69.32±1.39 & 73.20±0.55 & \textbf{81.99±1.64} \\
\hline
\multirow{4}{*}{\makecell{Wiki-CS \\ 10ways}} 
            & 0   & 50.84±2.05 & 45.84±0.87 & 49.25±2.31 & \textbf{55.48±1.05} \\
            & 5  & 50.80±2.28 & 59.05±1.01 & 60.20±1.09 & \textbf{62.24±2.18} \\
            & 20 & 51.18±1.88 & 60.92±0.85 & 62.60±1.41 & \textbf{68.20±2.40} \\
            & 50 & 50.34±1.77 & 59.25±1.31 & 65.80±1.33 & \textbf{75.81±2.34} \\
\hline
\multirow{4}{*}{\makecell{Products \\ 5ways}} 
            & 0   & 52.14±2.19 & 44.56±0.63 & 46.44±1.93 & \textbf{57.55±1.39} \\
            & 5  & 52.09±1.87 & 45.31±1.22 & 58.40±0.77 & \textbf{69.67±1.64} \\
            & 20 & 52.49±2.37 & 60.65±0.65 & 69.30±1.34 & \textbf{72.61±2.19} \\
            & 50 & 52.94±1.66 & 74.89±0.83 & 73.60±1.38 & \textbf{84.82±1.58} \\
\hline
\multirow{4}{*}{\makecell{Products \\ 10ways}} 
                & 0   & 37.43±2.20 & 42.77±0.77 & 41.35±2.20 & \textbf{44.80±1.20} \\
                & 5  & 37.87±2.41 & 54.29±1.26 & 56.20±1.23 & \textbf{58.30±2.17} \\
                & 20 & 37.18±2.39 & 60.96±1.05 & 56.60±1.20 & \textbf{64.85±1.86} \\
                & 50 & 37.35±1.54 & 64.62±0.62 & 59.02±0.61 & \textbf{75.81±1.52} \\     
\hline
\end{tabular}
}
\label{ablation_component}
}
\noindent\textbf{Ablation Study for Backbone Component.}
Secondly, we introduce several ablated models by removing different components: ``w/o Gates Module'' removes the gating function $t_{\boldsymbol{\phi}}$ and directly fine-tunes GNN encoder and hop encodings; ``w/o Hop Encoding'' means using zero vectors to replace the hop encodings $\{\boldsymbol{e}_i\}_{i=0}^{\lambda}$; ``w/o Task-Related Mask'' uses a random vector to initialize the Task-Related Mask $\boldsymbol{m}_{\text{rel}}$ instead of task texts. From the results in Table~\ref{ablation_component}, we can observe: (1) All components within the backbone are useful and necessary. Removing any part significantly degrades model performance. Specifically, removing the gates/hop encodings/task-related mask results in an average performance drop of 22.61\%/10.41\%/10.62\%, respectively. (2) Removing the Task-Related Mask leads to the largest drop in zero-shot performance, averaging 11.67\%. This drop is reasonable due to the absence of task-related information in node tokens.
{
\renewcommand{\arraystretch}{1.09}
\setlength{\tabcolsep}{1pt}
\small
\captionof{table}{Ablation for pre-training tasks.}
\resizebox{0.48\textwidth}{!}{
\begin{tabular}{@{}c|c|c|>{\columncolor{gray!30}}c|c|>{\columncolor{gray!30}}c|c@{}}
\Xhline{1.5pt}
Dataset     & Shots    & \makecell{w/o \\ Pre-training} & \makecell{Self-supervised \\ (ArXiv)} & \makecell{Self-super \\ + supervised \\ (ArXiv)} & \makecell{Self-super \\ + supervised \\(PubMed)} & \makecell{\modelname} \\ 
\hline
\multirow{4}{*}{\makecell{Cora \\ 5ways}}
            & 0  & 0          & 38.20±1.11 & 42.75±0.73 & 39.06±1.09 & \textbf{66.11±1.32} \\
            & 5  & 0          & 60.22±1.18 & 72.82±0.83 & 67.83±0.78 & \textbf{74.54±1.61} \\
            & 20 & 0          & 64.99±1.09 & 74.12±1.16 & 69.51±1.41 & \textbf{80.56±2.11} \\
            & 50 & 7.66±1.42  & 66.68±0.66 & 76.57±0.66 & 70.18±1.08 & \textbf{88.96±1.93} \\
\hline
\multirow{4}{*}{\makecell{Cora \\ 10ways}} 
            & 0  & 0          & 25.48±0.92 & 40.33±0.68 & 32.18±1.15 & \textbf{54.01±1.09} \\
            & 5  & 0.17±0.09  & 40.17±0.78 & 52.77±0.79 & 42.68±0.69 & \textbf{58.61±1.92} \\
            & 20 & 6.70±0.21  & 44.27±0.78 & 57.96±0.99 & 53.05±0.97 & \textbf{69.04±1.69} \\
            & 50 & 23.11±1.06 & 48.35±1.27 & 64.38±1.46 & 56.74±0.79 & \textbf{79.71±1.84} \\
\hline
\multirow{4}{*}{\makecell{Wiki-CS \\ 5ways}} 
            & 0  & 0          & 35.31±1.25 & 38.84±0.51 & 40.28±0.87 & \textbf{61.74±1.27} \\
            & 5  & 0          & 57.81±0.79 & 70.31±1.43 & 64.60±0.52 & \textbf{70.18±2.07} \\
            & 20 & 5.06±0.78  & 61.34±1.25 & 71.35±0.60 & 68.85±1.22 & \textbf{76.88±2.30} \\
            & 50 & 16.40±0.43 & 64.07±0.97 & 73.26±0.58 & 73.21±0.98 & \textbf{81.99±1.64} \\
\hline
\multirow{4}{*}{\makecell{Wiki-CS \\ 10ways}} 
            & 0  & 0          & 28.53±0.43 & 40.88±0.96 & 44.29±1.53 & \textbf{55.48±1.05} \\
            & 5  & 0          & 39.62±1.39 & 52.94±0.52 & 57.06±0.54 & \textbf{62.24±2.18} \\
            & 20 & 5.65±1.50  & 49.77±0.65 & 61.97±0.77 & 63.82±1.49 & \textbf{68.20±2.40} \\
            & 50 & 19.22±1.19 & 55.81±0.59 & 64.95±0.64 & 69.24±1.37 & \textbf{75.81±2.34} \\
\hline
\multirow{4}{*}{\makecell{Products \\ 5ways}} 
            & 0  & 0          & 47.99±0.81 & 54.52±0.54 & 51.37±0.24 & \textbf{57.55±1.39} \\
            & 5  & 0          & 61.98±1.13 & 60.31±0.25 & 58.28±0.15 & \textbf{69.67±1.64} \\
            & 20 & 0.17±0.03  & 65.13±0.07 & 71.34±0.54 & 71.04±1.37 & \textbf{72.61±2.19} \\
            & 50 & 2.41±0.56  & 72.15±0.57 & 78.94±0.51 & 80.05±1.45 & \textbf{84.82±1.58} \\
\hline
\multirow{4}{*}{\makecell{Products \\ 10ways}} 
            & 0  & 0          & 37.90±1.44 & 40.66±0.81 & 39.78±0.58 & \textbf{44.80±1.20} \\
            & 5  & 0          & 51.29±1.36 & 54.47±1.09 & 52.53±1.46 & \textbf{58.30±2.17} \\
            & 20 & 1.44±0.67  & 56.37±1.30 & 60.27±0.12 & 59.69±0.51 & \textbf{64.85±1.86} \\
            & 50 & 1.34±0.90  & 60.24±0.59 & 72.14±1.21 & 68.70±1.45 & \textbf{75.81±1.52} \\     
\hline
\end{tabular}
}
\label{ablation_tasks}
}
\noindent\textbf{Ablation Study for Pre-training Tasks.} 
We further conduct ablation experiments to evaluate the impact of each pre-training task, with results shown in Table~\ref{ablation_tasks}. Here ``w/o pre-training'' indicates that we conduct adaptation and inference on the model without any pre-training; ``Self-supervised (ArXiv)'' denotes that we pre-train the model solely with self-supervised tasks on the ArXiv dataset; ``Self-super + supervised (ArXiv)'' and ``Self-super + supervised (PubMed)'' mean that we conduct pre-training on the ArXiv or PubMed datasets using both self-supervised and supervised tasks. The results show that: (1) Each task and dataset involved in our pre-training are effective. The model that is pre-trained with all three tasks and on two datasets achieves the best performance. Specifically, the model with full pre-training performs, on average, 64.6\% better than the one without any pre-training, 17.91\% better than the model pre-trained only with self-supervised tasks, 8.89\% better than the model pre-trained only on the ArXiv dataset, and 10.87\% better than the one pre-trained only on the PubMed dataset. (2) The model without any pre-training performs the worst, with an average accuracy of only 4.25\%, indicating that it is yet unable to effectively comprehend node token information. (3) The model pre-trained only with self-supervised tasks shows significant improvement, with an average accuracy of 50.97\%. However, its problem-solving capability remains insufficient, ranking second to last in overall performance. (4) The model pre-trained on the ArXiv dataset with both self-supervised and supervised tasks outperforms the model pre-trained on the PubMed dataset. This result is expected, as the ArXiv dataset is larger than the PubMed dataset, providing richer information and enabling the model to develop stronger problem-solving capabilities.

\begin{table}[ht]
\centering
\caption{Efficiency comparison between \modelname and GraphGPT.}
\label{efficiency}
\resizebox{0.48\textwidth}{!}{
\begin{tabular}{@{}c|c|cc|cc@{}}
\Xhline{1.5pt}
\multicolumn{2}{c|}{\multirow{2}{*}{\large \textbf{Stage}}}  & \multicolumn{2}{c|}{\thead{\modelname}} & \multicolumn{2}{c}{\thead{GraphGPT}} \\
\cline{3-6}
\multicolumn{2}{c|}{} & \thead{Time-cost} & \thead{Tuned Parameters} & \thead{Time-cost} & \thead{Tuned Parameters}\\
\hline
\multirow{3}{*}{\makecell{Stage 1 \\ Pre-training}} 
        & Self-supervised & 25:25:47 & 131,827,714 & 23:28:21 & 131,612,672\\
        & Supervised (ArXiv) & 13:07:05 & 131,827,714 & 10:35:00 & 131,612,672 \\
        & Supervised (PubMed) & 10:13:50 & 131,827,714 & 9:11:27 & 131,612,672 \\
\hline
\multirow{3}{*}{\makecell{Stage 2 \\ Adaptation}}
        & Supervised (5-shots) & 0:01:40 & 726,658 & - & - \\
        & Supervised (20-shots) & 0:06:40 & 726,658 & - & - \\
        & Supervised (50-shots) & 0:16:40 & 726,658 & - & - \\
\hline
\multirow{4}{*}{\makecell{Stage 3 \\ Inference}}
        & 0-shot & 0.47s/target & - & 1.07s/target & - \\
        & 5-shots & 0.52s/target & - & 5.24s/target & - \\
        & 20-shots & 0.52s/target & - & 35.84s/target & - \\
        & 50-shots & 0.51s/target & - & 54s/target & - \\
\bottomrule
\end{tabular}
}
\end{table}
\subsection{Efficiency (RQ3)}
To answer \textbf{RQ3} and validate the efficiency, we measure the time costs and the number of trainable parameters at different stages. The results are presented in Table~\ref{efficiency}. Here we present the costs associated with pre-training stage (self-supervised training, supervised training on the ArXiv dataset, and supervised training on the PubMed dataset), adaptation stage (different shot counts), and inference stage (different shot counts). We also compare with baseline method GraphGPT with in-context learning inference.
From the experimental results, we find that:
(1) During the pre-training phase, our costs are roughly equivalent to the baseline, which lacks the additional components featured in our backbone design. 
(2) The time cost of adaptation is significantly less than that of the pre-training phase, averaging only about 1/117.  At the same time, the number of parameters involved in tuning is also small, approximately 1/181 of that in the pre-training phase. This indicates that the cost introduced by our adaptation phase is acceptable, achieving our design goals of efficient and effective transfer. 
(3) During the inference phase, \modelname consistently demonstrates stable and efficient performance, without increased consumption as the number of example shots increases. In contrast, the baseline's inference efficiency declines significantly as the number of shots increases 
averaging approximately 47.6 times that of GraphLAMA. Considering that the input length for ICL increases with the number of examples, this result is expected. 

\subsection{Case Study (RQ4)}
\begin{figure}[t]
    \centering
    \includegraphics[width=\linewidth]{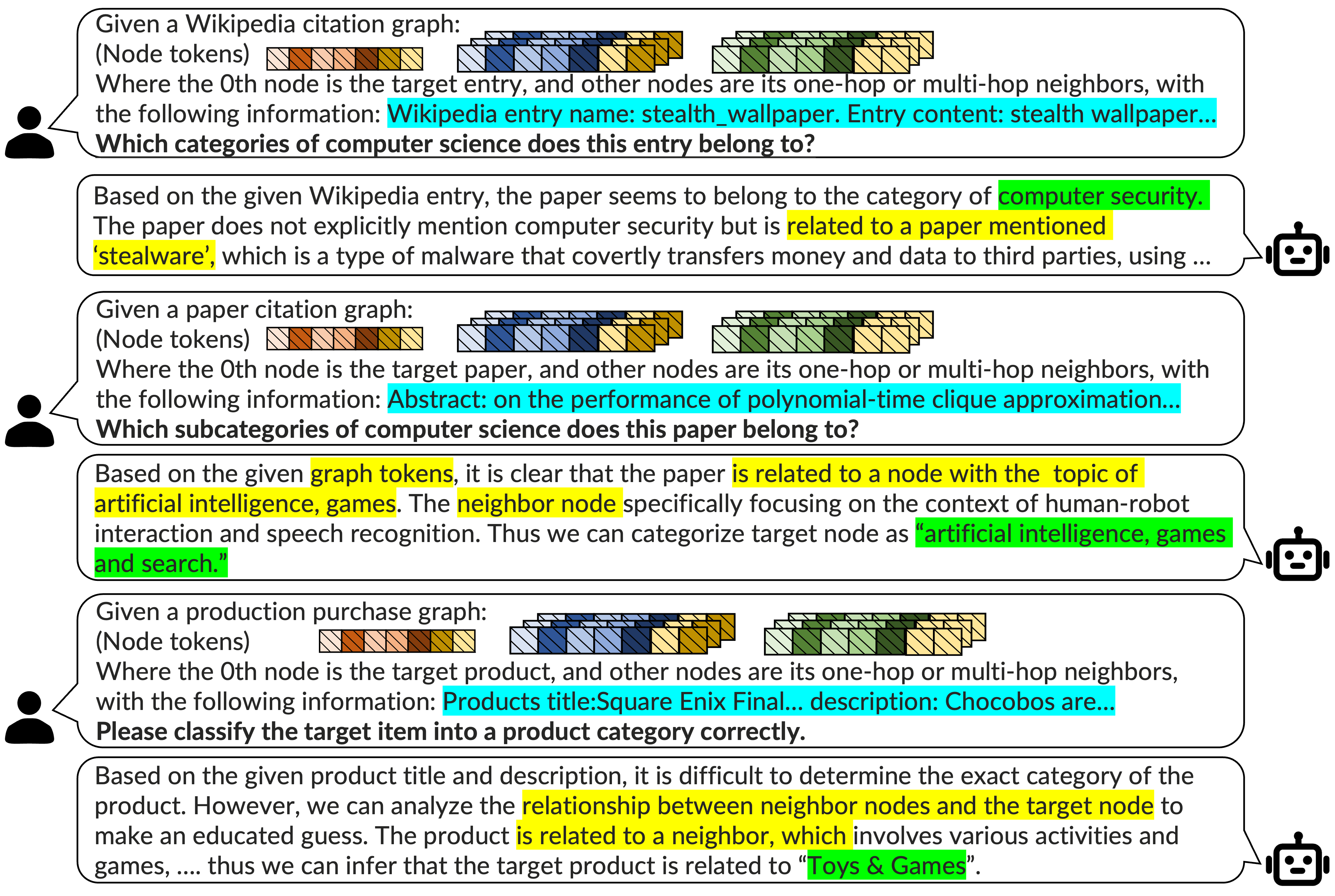}
    \caption{Case study of classification with explanations.}
    \label{fig:case}
\end{figure}

To further verify whether \modelname can effectively comprehend TAG and provide some interpretability for its inference, we select three cases from three different test datasets for demonstration. For each example, we display the input instructions along with the model's classification judgments and explanations, as shown in Figure~\ref{fig:case}. In the instructions, we highlight node content in blue, subgraph or neighbor info in yellow, and accuracy sections in green. The model effectively understands and uses information from both graph structure and text content, grasping the meaning of node tokens across different data domains.
When essential information is missing from the text content, the model can make accurate judgments based on neighbor relationships or subgraph configurations. 

\section{Conclusion}
In this paper, we discuss the limitations of existing adaptation paradigms for GLMs, and then propose to fine-tune GLMs for each target graph and task efficiently. The proposed \modelname method has well-designed backbone components and learning schemes, achieving better prediction accuracy and faster inference speed. Extensive results demonstrate that our proposed method is cost effective and our method design is valid. For future work, we will explore how the proposed \modelname performs on other graph learning tasks suffering from limited annotations, such as graph classification, and attempt to introduce additional new tasks during the pre-training stage to evaluate their effectiveness. Besides, explorations \modelname in more scenarios are also possible, such as molecular graphs or transportation networks, and mitigating the model’s hallucination issues is also worth exploring.

\section*{Acknowledgments}
This work was supported by the National Key Research and Development Program of China (No.2023YFC3303800). We thank the area chair and anonymous reviewers for their valuable feedback and constructive suggestions, which greatly helped improve the quality and clarity of this paper. 
\balance
\clearpage

\bibliographystyle{ACM-Reference-Format}
\bibliography{main/main}
\clearpage

\appendix

\section{Algorithm}
\label{pseudo-code}
\begin{algorithm}
	\renewcommand{\algorithmicrequire}{\textbf{Input:}}
	\renewcommand{\algorithmicensure}{\textbf{Output:}}
    \caption{Pre-training Stage of \modelname}
    \label{alg1}

    \begin{algorithmic}
        \STATE \textbf{Input:} Pre-training graph $\mathcal{G}_{\text{pre}} = \left(\mathcal{V}, \boldsymbol{A}, \boldsymbol{X}, \mathcal{S}\right)$,  GNN $g_{\theta}$, max hop number $\lambda$, hop encodings $\{\boldsymbol{e}_i\}_{i=0}^{\lambda}$, TRG and TIG $ t_{\phi} $, where $ \phi $ is divided into $ \phi_{\text{rel}} $,  $\phi_{\text{inv}} $, and $ \phi_{\text{agg}} $, $ \phi_{\text{rel}} $ includes $ \boldsymbol{W}_{\text{rel}} $, $ \phi_{\text{inv}} $ comprises $ \boldsymbol{m}_{\text{inv}} $ and $ \boldsymbol{W}_{\text{inv}} $, $ \phi_{\text{agg}} $ contains $ \boldsymbol{W}_{\text{agg}} $, projector $p_{\psi}$ , Large Language Model as encoder $\text{LLM}_{\omega}$, Task Text $T_{\text{task}}^{\text{pre}}$, text encoding function $f_{\alpha}$;
        \STATE \textbf{Output:} Well-trained transferable parameters $\theta$, $\phi$, $\psi$, $\{\boldsymbol{e}_i\}_{i=0}^{\lambda}$, $\alpha$;
        \STATE Initialize $\theta, \phi, \psi, \{\boldsymbol{e}_i\}_{i=0}^{\lambda}$;
        \FOR {each node $v \in \mathcal{V}$}
            \STATE Extract neighborhood subgraph $(\mathcal{N}(v), \boldsymbol{A}_{\mathcal{N}(v)}, \boldsymbol{X}_{\mathcal{N}(v)})$ from  $\mathcal{G}_{\text{pre}}$ of node $v$;
            \STATE $\boldsymbol{X}_{\mathcal{N}(v)}^{\text{GNN}} \gets g_{\theta}(\boldsymbol{A}_{\mathcal{N}(v)},\boldsymbol{X}_{\mathcal{N}(v)})$ by Eq.~(\ref{eq1});
            \STATE $\boldsymbol{m}_{\text{rel}} \gets f_{\alpha}({T}_{\text{task}}^{\text{pre}})$;
            \FOR{each node $u \in \mathcal{N}(v)$}
                \STATE $\boldsymbol{X}_u^{\text{Hop}} \gets [\boldsymbol{X}_u^{\text{GNN}}, \boldsymbol{e}_{\text{dist}(u, v)}]$ by Eq.~(\ref{eq2});
                \STATE $\boldsymbol{X}_u^{\text{Gate}} \gets t_{\phi}(\boldsymbol{X}_u^{\text{Hop}})$ by Eq.~(\ref{eq3});
                \STATE $\boldsymbol{X}_u^{\text{Proj}} \gets p_{\boldsymbol{\psi}}(\boldsymbol{X}_u^{\text{Gate}})$ by Eq.~(\ref{eq5});
                \STATE $\boldsymbol{C}_u \gets \text{LLM}_{{\omega}}(\mathcal{S}_u)$;
            \ENDFOR
            \STATE Update $\theta, \phi, \psi,\alpha, \{\boldsymbol{e}_i\}_{i=0}^{\lambda}$ by self-supervised loss in Eq.~(\ref{eq8}) and supervised cross-entropy loss;
        \ENDFOR
    \end{algorithmic}
\end{algorithm}

\begin{algorithm}
    \caption{Adaptation Stage of \modelname}
    \label{alg3}
    \begin{algorithmic}
        \STATE \textbf{Input:} Target graph $\mathcal{G}_{\text{target}} = \left(\mathcal{V}, \boldsymbol{A}, \boldsymbol{X}, \mathcal{S}\right)$, few-shot examples $\{(v_i, y_{v_i}\}_{i=1}^{K}$, GNN $g_{\theta}$, max hop number $\lambda$, hop encodings $\{\boldsymbol{e}_i\}_{i=0}^{\lambda}$, TRG and TIG $ t_{\phi} $, where $ \boldsymbol{\phi} $ is divided into $ \phi_{\text{rel}} $,  $\phi_{\text{inv}} $, and $ \phi_{\text{agg}} $, $ \phi_{\text{rel}} $ includes $ \boldsymbol{W}_{\text{rel}} $, $ \phi_{\text{inv}} $ comprises $ \boldsymbol{m}_{\text{inv}} $ and $ \boldsymbol{W}_{\text{inv}} $, $ \phi_{\text{agg}} $ contains $ \boldsymbol{W}_{\text{agg}} $, projector $p_{\psi}$, Large Language Model as decoder $\text{LLM}_{\omega}$, Task Text ${T}_{\text{task}}^{\text{target}}$, text encoding function $f_{\alpha}$;
        \STATE \textbf{Output:} Well tuned parameters for target task and data $\alpha, \phi_{\text{agg}}$;
        \STATE Initialize $\boldsymbol{m}_{\text{rel}} \gets f_{\alpha}({T}_{\text{task}}^{\text{target}})$
        \FOR {$i=1 \TO k$}
            \STATE Extract neighborhood subgraph $(\mathcal{N}(v), \boldsymbol{A}_{\mathcal{N}(v)}, \boldsymbol{X}_{\mathcal{N}(v)})$ from  $\mathcal{G}_{\text{target}}$ of node $v_i$;
            \STATE $\boldsymbol{X}_{\mathcal{N}(v)}^{\text{GNN}} \gets g_{\theta}(\boldsymbol{A}_{\mathcal{N}(v)},\boldsymbol{X}_{\mathcal{N}(v)})$ by Eqs.~(\ref{eq1});
            \FOR{each node $u \in \mathcal{N}(v)$}
                \STATE $\boldsymbol{X}_u^{\text{Hop}} \gets [\boldsymbol{X}_u^{\text{GNN}}, \boldsymbol{e}_{\text{dist}(u, v)}]$ by Eq.~(\ref{eq2});
                \STATE $\boldsymbol{X}_u^{\text{Gate}} \gets t_{\phi}(\boldsymbol{X}_u^{\text{Hop}})$ by Eq.~(\ref{eq3});
                \STATE $\boldsymbol{X}_u^{\text{Proj}} \gets p_{\psi}(\boldsymbol{X}_u^{\text{Gate}})$ by Eq.~(\ref{eq5});
            \ENDFOR
            \STATE $\hat{y}_{v_i} \gets LLM_{{\omega}}[\boldsymbol{X}_{\mathcal{N}(v)}^{\text{Proj}}, \mathcal{S}_v, {T}_{\text{task}}^{\text{target}}]$ by Eq.~(\ref{eq7});
            \STATE Update $\boldsymbol{m}_{\text{rel}}, \phi_{\text{agg}}, \phi_{\text{rel}}$ by supervised cross-entropy loss;
        \ENDFOR
    \end{algorithmic}
\end{algorithm}

\section{Dataset Details}
\label{datasets-details}
Arxiv \cite{mccallum2000automating} represents a directed graph depicting the citation network among computer science ArXiv papers. Each paper in the dataset is categorized into one of 40 specific research fields, with the classifications provided manually by the authors and verified by ArXiv moderators. The PubMed dataset \cite{he2023explanations} comprises 19,717 scientific articles related to diabetes, sourced from the PubMed database. These articles are systematically classified into three categories: Experimental Induced Diabetes, Type 1 Diabetes, and Type 2 Diabetes. Furthermore, the dataset features a citation network that includes 44,338 connections. The Cora dataset \cite{wen2023augmenting} contains 25,120 research papers, interconnected through a network of citations. This expanded version of the Cora dataset includes a total of 70 classes. Wiki-CS \cite{liu2023one} constitutes a network of internet links, with each node depicting a Wikipedia page and each edge denoting a hyperlink between pages. The ogbn-products dataset \cite{hu2020open} comprises an undirected and unweighted graph that models a network of co-purchased products on Amazon. In this graph, nodes symbolize products available on Amazon, and edges between two nodes signify that the corresponding products are frequently bought together. The sources of datasets are list as follows:
\begin{itemize}
    \item \textbf{ArXiv}: Graph and node content at \url{https://ogb.stanford.edu/docs/nodeprop/}
    \item \textbf{PubMed}: Graph and node content at \url{https://github.com/XiaoxinHe/TAPE}
    \item \textbf{Cora}: Graph and node content at \url{https://github.com/WenZhihao666/G2P2}
    \item \textbf{Wiki-CS}: Graph and node content at \url{https://github.com/LechengKong/OneForAll}
    \item \textbf{Products}: We get graph data from \url{https://ogb.stanford.edu/docs/nodeprop/} and get node content from \url{https://github.com/XiaoxinHe/TAPE}
\end{itemize}

\section{Baselines}
\label{baseline-details}
For few-shot classification task, baseline methods can be divided into two groups. (1) The first group employs discriminative learning approaches and uses GNN as predictor:GraphMAE \cite{hou2022graphmae} employs self-supervised learning method and designed for cross-domain purpose; All-In-One \cite{sun2023all} and GraphPrompt \cite{liu2023graphprompt} adopt graph prompt methods and are capable of few-shot learning; Prodigy \cite{huang2024prodigy} and OneForAll (OFA) \cite{liu2023one} enable in-context learning on graph data; G2P2 \cite{wen2023augmenting} utilizes natural language prompt methods and integrates soft-prompt techniques with graph structures.
For the baselines other than GraphMAE, All-In-One, and GraphPrompt, we employed the standard word2vec method to encode textual information.
These approaches are capable of few-shot classification but not summary generation. (2) The second group consists of methods that use LLMs to generate predictions, including Vicuna-7B-v1.5, NLGraph \cite{wang2024can}, LLaGA \cite{chen2024llaga}, G-Retriever \cite{He2024GRetrieverRG} and GraphGPT \cite{tang2023graphgpt}. Typically, we describe graph structure as natural language to be the input of Vicuna and NLGraph, along with node's text content. The inputs of GraphGPT include graph tokens for structure information and textual features. 
We incorporate few-shot examples into the instruction prompts during the inference phase for (ICL) tagged models, and use few-shot data to adjust the parameters of the (Tuned) tagged models, as suggested by their original papers.

For zero-shot classification experiments, we compare three groups of methods as baselines. (1) The first group consists of supervised GNNs, including GCN \cite{hamilton2017inductive}, GAT \cite{velickovic2017graph} and Graph-SAGE \cite{kipf2016semi} which are trained and evaluated on test datasets. (2) The second group comprises standalone LLMs, specifically Llama2-7B and Vicuna-7B-v1.5, which only use text contents of nodes for classification predictions. (3) The third group includes methods that integrate GNNs with LLMs such as GraphGPT, OneForAll, GIMLET \cite{zhao2024gimlet}, GraphTranslator~\cite{zhang2024graphtranslator} and LLaGA. Models in the third group including our \modelname are trained on the pre-training datasets and then directly evaluated on the test datasets.

For summary generation tasks, we select Llama2-7B and Vicuna-7B-v1.5 as text-only baselines. Additionally, we employ GraphTranslator as a baseline that utilizes both graph structure and textual information. We ignore GraphGPT in this task, since it fails to follow the summary instruction. 

\section{Implementation Details}
\label{implementation-details}
For each graph used in pre-training and testing, we get 128-dimensional node features through encoding node content with word2vec \cite{mikolov2013efficient}. The GNN used in Eq.~(\ref{eq1}) is realized with a three-layer Graph Transformer \cite{yun2019graph} and the hidden dimension is set at 128. For hop encoding, we set it as a fixed-length vector of size 4. During all stages, we set $\lambda$ as 2 to extract the two-hop subgraph of the target node, and set the max number of neighbors as 100. We use 132-dimensional vectors for $\boldsymbol{m}_{\text{inv}}$ and $\boldsymbol{m}_{\text{rel}}$ used in Eq.~(\ref{eq3}). For $f_{\alpha}$, we utilize the ``all-MiniLM-L6-v2'' version of Sentence-bert. For projector, we implement a fully connected layer for $p_{\psi}$ used in Eq.~(\ref{eq5}). We utilize Vicuna-7B-v1.5 as the LLM in our backbone architecture, which features a hidden dimension of 4096. For Eq.~(\ref{eq8}), we use cosine similarity normalization for the term $\cos(\hat{\boldsymbol{X}}_u, \hat{\boldsymbol{C}}_u)$ and magnitude normalization for the Euclidean distance component $\|\hat{\boldsymbol{X}}_u - \hat{\boldsymbol{C}}_u\|^2$.
We employ the Adam optimizer with a learning rate of 1e-6 for both the pre-training and adaptation stages. The model is pre-trained for up to three epochs and undergo adaptation for a maximum of two epochs. Considering the computational cost in pre-training, we simply follow the experience in GraphGPT for hyper-parameter selection, and use the above setting for all datasets and tasks. All experiments are conducted on an A800 GPU with 80GB of memory.


\section{Comparison with Directly Scaling GLMs}
\label{scaling}
We equip GraphGPT with Vicuna-13B-v1.5~\cite{chiang2023vicuna} and Llama-8B-v3~\cite{touvron2023llama} for comparison, with results shown in Table~\ref{graphgpt_vs_original}. Notably, GraphLAMA uses Vicuna-7B-v1.5 as its LLM. As the results indicate, our method consistently outperforms all other configurations across the board. The lack of training specifically tailored to graph data may explain why simply scaling up the LLM’s parameters and training corpus fails to yield better performance.
{
\renewcommand{\arraystretch}{1.05}
\small
\setlength{\tabcolsep}{2pt}
\captionof{table}{Comparison between GraphGPT variants and \modelname.}
\resizebox{0.48\textwidth}{!}{
\begin{tabular}{@{}c|c|>{\columncolor{gray!30}}c|c|c@{}}
\Xhline{1.5pt}
Dataset     & Shots & \makecell{GraphGPT with \\ Vicuna-13B-v1.5 \\ (ICL)} & \makecell{GraphGPT with \\ Llama-8B-v3 \\ (ICL)} & \makecell{\modelname } \\ 
\hline
\multirow{3}{*}{\makecell{Cora \\ 5ways}}
            & 5  & 51.08±1.21 & 51.60±2.22 & \textbf{74.54±1.61} \\
            & 20 & 50.84±1.45 & 51.75±2.20 & \textbf{80.56±2.11} \\
            & 50 & 53.73±1.68 & 52.86±2.22 & \textbf{88.96±1.93} \\
\hline
\multirow{3}{*}{\makecell{Cora \\ 10ways}}
            & 5  & 39.95±1.36 & 39.47±1.34 & \textbf{58.61±1.92} \\
            & 20 & 39.77±1.69 & 39.16±1.73 & \textbf{69.04±1.69} \\
            & 50 & 40.25±1.10 & 40.59±1.55 & \textbf{79.71±1.84} \\
\hline
\multirow{3}{*}{\makecell{Wiki-CS \\ 5ways}}
            & 5  & 55.34±1.72 & 56.08±1.06 & \textbf{70.18±2.07} \\
            & 20 & 55.74±1.60 & 56.47±1.30 & \textbf{76.88±2.30} \\
            & 50 & 57.33±2.26 & 57.72±2.10 & \textbf{81.99±1.64} \\
\hline
\multirow{3}{*}{\makecell{Wiki-CS \\ 10ways}}
            & 5  & 50.53±2.10 & 49.21±1.36 & \textbf{62.24±2.18} \\
            & 20 & 49.52±2.30 & 49.74±1.09 & \textbf{68.20±2.40} \\
            & 50 & 50.70±1.17 & 49.92±1.51 & \textbf{75.81±2.34} \\
\hline
\multirow{3}{*}{\makecell{Products \\ 5ways}}
            & 5  & 51.61±1.48 & 50.27±1.75 & \textbf{69.67±1.64} \\
            & 20 & 51.96±1.43 & 50.33±2.38 & \textbf{72.61±2.19} \\
            & 50 & 51.83±2.05 & 50.09±2.20 & \textbf{84.82±1.58} \\
\hline
\multirow{3}{*}{\makecell{Products \\ 10ways}}
            & 5  & 36.88±2.45 & 36.72±2.28 & \textbf{58.30±2.17} \\
            & 20 & 36.29±1.43 & 36.38±1.77 & \textbf{64.85±1.86} \\
            & 50 & 37.51±1.76 & 36.71±1.08 & \textbf{75.81±1.52} \\     
\hline
\end{tabular}
}
\label{graphgpt_vs_original}
}

\end{document}